\PassOptionsToPackage{numbers,sort&compress}{natbib}
\documentclass{article}

\usepackage[final]{neurips_2025}

\usepackage[utf8]{inputenc}
\usepackage[T1]{fontenc}

\usepackage{url}
\usepackage{booktabs}
\usepackage{amsfonts}
\usepackage{nicefrac}
\usepackage{microtype}
\usepackage{xcolor}
\usepackage{graphicx}
\usepackage{multirow}
\usepackage{makecell}
\usepackage{amsmath}
\usepackage{array}
\usepackage{float}
\newcolumntype{C}[1]{>{\centering\arraybackslash}m{#1}}
\title{
SkyCap: Bitemporal VHR Optical–SAR Quartets for Amplitude Change Detection and Foundation-Model Evaluation\\[0.2em]
\large\textnormal{Accepted at the Workshop\\
\textit{Advances in Representation Learning for Earth Observation (REO)}\\
EurIPS 2025, December 7, 2025}
}

\author{
  Paul Weinmann \quad
  Ferdinand Schenck \quad
  Martin \v{S}iklar \\
  LiveEO GmbH \\
  Berlin, Germany \\
  \texttt{\{paul.weinmann, ferdinand.schenck, martin.siklar\}@live-eo.com}
}

\begin{document}

\maketitle

\begin{abstract}
Change detection for linear infrastructure monitoring requires reliable high-resolution data and regular acquisition cadence. Optical very-high-resolution (VHR) imagery is interpretable and straightforward to label, but clouds break this cadence.
Synthetic Aperture Radar (SAR) enables all-weather acquisitions, yet is difficult to annotate. We introduce \textbf{SkyCap}, a bitemporal VHR optical–SAR dataset constructed by archive matching and co-registration of (optical) \textbf{Sky}Sat and \textbf{Cap}ella Space (SAR) scenes. We utilize optical-to-SAR label transfer to obtain SAR amplitude change detection (ACD) labels without requiring SAR-expert annotations.
We perform continued pretraining of SARATR-X on our SAR data and benchmark the resulting SAR-specific foundation models (FMs) together with SARATR-X against optical FMs on SkyCap under different preprocessing choices. Among evaluated models, MTP(ViT-B+RVSA), an optical FM, with dB+Z-score preprocessing attains the best result (F1\textsubscript{c} = 45.06), outperforming SAR-specific FMs further pretrained directly on Capella data. We observe strong sensitivity to preprocessing alignment with pretraining statistics, and the ranking of optical models on optical change detection does not transfer one-to-one to SAR ACD. To our knowledge, this is the first evaluation of foundation models on VHR SAR ACD.

\end{abstract}

\section{Introduction}\label{sec:intro}
Monitoring linear infrastructure such as pipelines, power lines, and railways requires reliable high-resolution change detection with consistent revisit intervals. For pipelines, this cadence is often mandated by regulation~\citep{phmsa_regulations}.
Due to its high interpretability, optical very high-resolution (VHR) imagery supports both reliable human labeling and effective machine learning analysis. However, it is fundamentally constrained by cloud cover, disrupting targeted revisit intervals~\citep{cloud_cover}.\\
Synthetic Aperture Radar (SAR), by contrast, enables imaging regardless of weather or daylight conditions, making it a robust alternative for consistent monitoring. To capture the small-scale changes relevant to infrastructure monitoring, we focus specifically on VHR SAR data. However, SAR interpretation poses significant challenges: complex backscatter patterns, speckle noise, and geometric distortions make direct annotation difficult, especially given limited expert capacity. These limitations motivate our exploration of SAR amplitude change detection (ACD) as a core component of infrastructure monitoring, aiming to combine SAR's acquisition reliability with foundation model (FM)-based CD methods. \\
Foundation models (FMs) have improved transfer learning in both vision~\citep{mae} and remote sensing~\citep{sat_mae}. SAR-specific pretraining approaches such as SARATR-X adapt Masked Autoencoder pretraining to improve stability under speckle noise~\cite{saratrx}, but are evaluated primarily on few-shot object detection. It remains unclear how well such models transfer to the VHR SAR ACD task for infrastructure monitoring, and how they compare to optical FMs applied to SAR with appropriate preprocessing. \\
To address SAR labeling challenges, we construct \textit{SkyCap}, a bitemporal optical–SAR dataset by cross-referencing Capella X-band Spotlight SAR with Planet SkySat optical pairs and transferring change labels from optical to co-registered SAR. \\
\textbf{We study three questions:}
(1) How can we obtain reliable SAR change labels without SAR-expert annotation?
(2) How do SAR pretrained foundation models compare to optical foundation models transferred to SAR for change detection?
(3) Which preprocessing enables the best cross-modal transfer, and does the relative ranking of optical FMs translate to SAR? \\
\textbf{Our contributions are:}
(i) a practical optical-to-SAR label transfer pipeline for VHR change detection;
(ii) a controlled comparison of continued SAR pretraining on Capella and ALOS-2 SAR data against transferring optical FMs;
(iii) an empirical result that optical FMs outperform SAR-specific pretraining on Capella Space Spotlight data, along with an analysis of preprocessing influence.

\section{Related Work}\label{sec:related}
\textbf{SAR Amplitude Change Detection (ACD).}
Prior work on VHR amplitude-only SAR CD has focused on TerraSAR-X and COSMO-SkyMed under repeat-pass, matched-geometry pairs, with mostly qualitative case studies or limited metrics on small datasets~\citep{lanfri2013change, boldt2012, weihing2010tsx_change}.
\textbf{SAR Foundation Models.} SARATR-X~\citep{saratrx}, SAR-JEPA~\citep{li2024} and MSFA~\citep{sardet100k} explore improved representations for the Masked Autoencoder reconstruction target to improve the stability of MAE pretraining on speckle-afflicted SAR data. Multimodal approaches such as TerraMind~\citep{terramind} and DOFA~\citep{dofa} simultaneously pretrain on optical and SAR data. However, these multimodal approaches primarily target medium-resolution dual-polarization, C-band Sentinel-1 data rather than commercial VHR SAR (single-pol, X-band).
\textbf{Optical Foundation Models for Remote Sensing.} MTP~\citep{mtp} leverages multi-task pretraining on diverse annotated datasets, while USat~\citep{usat} integrates multiple spectral bands and spatial resolutions. DINOv3~\citep{dinov3} introduces a general domain-agnostic training approach for optical data and the results on remote sensing data support the view that domain-agnostic pretraining can transfer effectively to specialized downstream domains~\citep{self-sup_sat}. Despite their success on optical data, systematic evaluation of their transfer to VHR SAR ACD remains unexplored.

\section{Methodology}\label{sec:method}

\label{subsec:dataset}
\textbf{SkyCap Dataset Creation} We build incidental, bitemporal optical–SAR quartets by archive-matching Capella Space Spotlight SAR (submeter GSD, X-band, HH polarization) and Planet SkySat optical scenes (0.5m GSD, RGB+NIR), then co-registering all scenes.

We selected locations with a high probability of human-induced change, i.e., near human settlements and substantial temporal separation. For more details, see Appendix~\ref{app:dataset}.
After location-based deduplication, we obtained 19 scene \textit{quartets} (i.e. time step 1 and 2 each consist of a SkySat and Capella scene, in total four involved scenes) covering Eastern Europe, the Middle East and most of Asia across tropical, desert, and temperate biomes. This results in 3,484 annotated image-pair samples with changes.
\textbf{Annotation Strategy} All change annotations were created on interpretable optical image-pairs by an experienced annotation team, then transferred to co-registered SAR pairs.

\textbf{Continued Pretraining.}
We extend the SARATR-X pretraining pipeline by continuing pretraining on our SAR data using model weights obtained upon request from the authors, following their MSGF-based objective and configuration. SARATR-X is based on HiViT-B~\citep{hivit} with ImageNet initialization and modifies the MAE objective~\citep{mae} by replacing backscatter intensities with Multi-Scale Gradient Features (MSGF). MSGF serve as a denoised representation of image content, reducing sensitivity to multiplicative speckle noise and stabilizing the pretraining objective.
This yields three model variants trained for $\sim$73\% of the SARATR-X training schedule:
\textbf{CapellaX} was trained on 136k Capella X-band images.
\textbf{ALOS-X} used 254k ALOS-2 L-band images.
\textbf{CapALOS-X} combined both datasets in a 50/50 split.
ALOS-X and CapALOS-X serve to evaluate cross-sensor transfer from ALOS-2/PALSAR-2 (10m; L-band) to Capella Space Spotlight (0.5m; X-band). We analyze how pretraining on these distinct SAR sources affects downstream performance on Capella data.

\textbf{Change Detection Task} We evaluate two tasks separately on SkyCap: a) Optical Change Detection on the SkySat pair from the SkyCap quartet utilizing the optical-derived labels, referred to as SkyCap optical b) SAR Amplitude Change Detection on the Capella pair from the same SkyCap quartets utilizing the same optical-derived labels as in a), referred to as SkyCap SAR. This allows us to directly compare SAR ACD with optical CD.

\textbf{Change Detection Architecture.}

We employ a simple middle-fusion siamese architecture with the respective investigated encoder as the backbone, an absolute difference neck, and a U-Net~\citep{unet} decoder. This matches the model architecture utilized for Change Detection in MTP~\citep{mtp}.

\textbf{SAR Preprocessing.} We evaluate three preprocessing configurations: (1) \textit{linear}: clip amplitudes to 0.5-99.5 percentiles of the training distribution and scale to [0,1], this aims to follow the preprocessing employed in SARATR-X; (2) \textit{linear+Z-score }: linear preprocessing followed by Z-score normalization; (3) \textit{dB+Z-score }: converts amplitudes to decibels, followed by applying Z-score normalization. The decibel scaling transforms the right-skewed gamma-distributed SAR intensities into an approximately normal distribution, more closely matching the distribution of natural optical images. SARATR-X uses linear preprocessing, and its multi-scale gradient features are not directly compatible with dB inputs, motivating our linear-input setting for SAR-pretrained models.

\textbf{Training.}
We evaluate six encoders in total, three of which are optical, i.e. HiViT~\citep{hivit}, MTP(ViT-B+RVSA)\citep{mtp}, and DINOv3~\citep{dinov3}, and three SAR pretrained models SARATR-X~\citep{saratrx}, CapellaX, and CapALOS-X. CapellaX and CapALOS-X both follow the training approach of SARATR-X and were further pretrained on Capella Space sensor data. We limit the evaluation on optical data to optical models. All evaluated encoders are of the Base size ($\sim$90M parameters) for a fair comparison. For more details, see Appendix~\ref{app:training}.

\section{Results}\label{sec:results}
We report \textbf{F1\textsubscript{c}}, \textbf{IoU\textsubscript{c}}, \textbf{Prec\textsubscript{c}} (precision), \textbf{Recall\textsubscript{c}} for the \emph{change} class only multiplied by 100. All absolute differences are expressed in percentage points (pp).
\begin{table*}[t]
\centering
\small
\setlength{\tabcolsep}{3pt}
\renewcommand{\arraystretch}{1.05}

\begin{minipage}[t]{0.5\textwidth}
\centering
\caption{SkyCap SAR (Capella X-band). Change-class metrics (×100, higher is better). Best per FM underlined, best overall in bold.}
\label{tab:skycap_results}
\resizebox{\linewidth}{!}{
\begin{tabular}{C{2cm} C{2cm} c c c c}
\toprule
\textbf{Encoder} & \textbf{Preprocessing} & \textbf{$\text{F1}_c$} & \textbf{$\text{IoU}_c$} & \textbf{$\text{Prec}_c$} & \textbf{$\text{Recall}_c$} \\
\midrule
\multirow{3}{*}{\makecell{HiViT \\ (optical)}}
 & linear         & 41.63 & 28.44 & \underline{39.39} & 46.86 \\
 & linear+Z-score  & 41.71 & 28.60 & 38.63 & 48.41 \\
 & dB+Z-score      & \underline{42.11} & \underline{29.04} & 39.15 & \underline{48.78} \\
\midrule
\multirow{2}{*}{\makecell{SARATR-X}}
 & linear         & \underline{40.03} & \underline{27.11} & \underline{36.99} & \underline{46.19} \\
 & dB+Z-score      & 34.97 & 23.14 & 32.04 & 42.22 \\
\midrule
\multirow{2}{*}{\makecell{CapellaX}}
 & linear         & \underline{42.06} & \underline{28.90} & \underline{38.86} & 48.44 \\
 & dB+Z-score      & 38.76 & 26.15 & 32.04 & \underline{48.95} \\
\midrule
\multirow{2}{*}{\makecell{CapALOS-X}}
 & linear         & \underline{44.35} & \underline{30.87} & \underline{41.81} & \underline{49.94} \\
 & dB+Z-score      & 39.70 & 27.04 & 36.53 & 46.13 \\
\midrule
\multirow{3}{*}{\makecell{MTP \\ ViT+RVSA \\ (optical)}}
 & linear         & 42.20 & 29.35 & 40.53 & 48.93 \\
 & linear+Z-score  & 44.52 & 31.12 & 40.60 & \underline{52.22} \\
 & dB+Z-score      & \textbf{\underline{45.06}} & \textbf{\underline{31.68}} & \underline{41.73} & 51.51 \\
\midrule
\multirow{3}{*}{\makecell{DINOv3 \\ ViT \\ (optical)}}
 & linear         & 41.60 & 28.80 & \underline{38.80} & 48.63 \\
 & linear+Z-score  & 41.20 & 28.41 & 37.42 & 49.37 \\
 & dB+Z-score      & \underline{42.40} & \underline{29.18} & 36.84 & \textbf{\underline{53.52}} \\
\midrule
\multirow{3}{*}{\makecell{DINOv3 \\ ConvNeXt \\ (optical)}}
 & linear         & 43.53 & 30.47 & 40.80 & 49.27 \\
 & linear+Z-score  & \underline{44.25} & 31.03 & 41.92 & \underline{49.29} \\
 & dB+Z-score      & 44.18 & 31.02 & \textbf{\underline{45.57}} & 44.96 \\
\bottomrule
\end{tabular}}
\end{minipage}
\hspace{0.02\textwidth}
\begin{minipage}[t]{0.45\textwidth}
\centering
\caption{Results on SkyCap optical (SkySat). Change-class metrics (×100, higher is better). Best overall in bold.}
\label{tab:skycap_optical_results}
\resizebox{\linewidth}{!}{
\begin{tabular}{lcccc}
\toprule
\textbf{Encoder} & \textbf{F1} & \textbf{IoU} & \textbf{Prec.} & \textbf{Recall} \\
\midrule
HiViT (optical) & 63.31 & 48.18 & 60.15 & 68.67 \\
MTP(ViT+RVSA)& 60.86 & 45.44 & 57.05 & 66.46 \\
DINOv3 ViT & 66.07 & 50.65 & 63.40 & \textbf{69.96} \\
DINOv3 ConvNeXt & \textbf{68.18} & \textbf{53.25} & \textbf{67.25} & 69.82 \\
\bottomrule
\end{tabular}}
\end{minipage}
\vspace{-0.6em}
\end{table*}

\subsection{Change Detection Results}

\textbf{SkyCap (Capella X-band).} Table~\ref{tab:skycap_results} reports test results. With \emph{linear} inputs, all optical models outperform SARATR-X (F1 = 40.03). Continued pretraining on Capella improves performance (CapellaX F1 = 42.06, +0.43 pp vs HiViT, +2.03 pp vs SARATR-X), and CapALOS-X is the strongest SAR-pretrained model (F1 = 44.35, +2.72 pp vs HiViT, +4.32 pp vs SARATR-X). MTP(ViT-B+RVSA) with \emph{dB+Z-score} achieves the best overall result (F1 = 45.06). Preprocessing matters: \emph{dB+Z-score} improves optical Transformers by +2.32 to +2.86 pp but reduces SAR-pretrained models by -1.27 to -5.06 pp. For DINOv3 ConvNeXt-B, \emph{linear+Z-score} slightly exceeds \emph{dB+Z-score}.

\textbf{SkySat (optical) data.}
For comparison, we report results on the optical parts of the quartets from which the annotations were obtained in Table~\ref{tab:skycap_optical_results}. The models were trained in a similar fashion to the SAR case, but no continued pretraining was performed. The best results were achieved by the ConvNeXt version of DINOv3~\citep{dinov3} with an F1 score of 68.18.

\section{Discussion and Conclusion}\label{sec:conclusion}
\textbf{Q1.} We obtain reliable SAR change labels without SAR-expert annotation by transferring labels from interpretable optical imagery to co-registered SAR within our multimodal quartets, producing SAR data to train binary ACD models and systematically evaluate the impact of foundation-model backbones (encoders).\\
\textbf{Q2.} On high-resolution Capella X-band, optical foundation models with \emph{dB+Z-score} achieve the best performance (best: MTP, F1 = 45.06), outperforming SAR-specific approaches including continued pretraining on target-sensor data. This is surprising as the SAR ACD data is in distribution for the SAR FMs and out of distribution for the optical FMs.
We hypothesize that optical FMs outperform the evaluated SAR-specific models in our setting because dB+Z-score preprocessing reshapes the roughly gamma-distributed SAR amplitudes into an approximately normal distribution closer to natural images and  enhances low-intensity structural contrast, whereas the SARATR-X-style pretraining reconstruction target (MSGF) emphasizes bright scatterers, which is helpful for Automatic Target Recognition but less well aligned with the subtle, low-backscatter changes that dominate SkyCap. \\
\textbf{Q3.} Preprocessing alignment is central. Models perform best when evaluation inputs match their pretraining statistics. Optical FMs gain under \emph{dB+Z-score}, while SAR-pretrained models trained on linear amplitudes lose accuracy under dB inputs. The ranking from optical CD does not transfer one-to-one to SAR ACD: DINOv3 ConvNeXt leads on optical, MTP(ViT+RVSA) leads on SAR. We hypothesize that architectural factors, for example MTP’s Rotated Variable Sized Attention~\citep{rvsa}, contribute more to transfer performance than differences in optical feature quality alone.
The gap between the best optical result (F1 = 68.18) and the best SAR result (F1\textsubscript{c} = 45.06) is 23.12\,pp, underscoring a persistent modality gap between VHR optical and SAR amplitude change detection.
These findings indicate that careful input transformation can substitute for costly SAR-specific pretraining in VHR SAR amplitude change detection, and that cross-modal label transfer is a practical path to create evaluation data at scale.

\subsection{Limitations}
\label{subsec:limitations}
To our knowledge, this is the first evaluation of foundation models for very high resolution (VHR) SAR amplitude change detection (ACD). The findings should therefore be viewed as indicative rather than conclusive. The dataset’s size and geographic coverage remain limited, preventing strong claims about statistical significance or global generalization. We rely on random patch-level splits instead of geographically disjoint splits, which constrains conclusions about spatial generalization. Despite manual refinement, small co-registration errors between optical and SAR imagery introduce residual geometric misalignments and label noise. Moreover, SAR backscatter often diverges from true object geometry, causing discrepancies between optical labels and SAR signal responses. Temporal offsets of up to five days between optical and SAR acquisitions may further introduce mismatched change labels, and not all optically visible changes necessarily yield measurable X-band backscatter differences.
\subsection{Future Work}

Key directions include:
(1) Investigating why optical foundation models transfer successfully despite fundamental physical differences between optical and SAR sensing;
(2) Further improving \textit{SkyCap} by (a) adapting optical-derived label masks to better match SAR backscatter responses, and (b) analyzing which object classes or change types are poorly represented or invisible in SAR backscatter and refining the label set accordingly;
(3) Exploring multimodal pretraining strategies that jointly learn from optical and SAR imagery to capture complementary information.

In conclusion, we demonstrate a path toward reliable all-weather change detection for infrastructure monitoring by combining optical supervision with SAR acquisition through multimodal dataset creation, and show that optical foundation models with appropriate preprocessing can outperform SAR-specific pretraining on very high-resolution data.

\section*{Acknowledgments}

This work was supported by the German Federal Ministry for Economic Affairs and Climate Action (grant 50EE2016).
We thank Capella Space and Planet Labs PBC for providing access to archival imagery.
Any opinions and conclusions are those of the authors and do not necessarily reflect the views of the data providers.

\bibliographystyle{plainnat}
\bibliography{references}

\appendix
\newpage

\section{Dataset Details}
\label{app:dataset}

\begin{table}[H]
\centering
\caption{SAR Imaging Geometry Constraints}
\begin{tabular}{ll}
\hline
\textbf{Parameter} & \textbf{Requirement} \\
\hline
Orbit & Identical \\
Orbit direction & Identical \\
Observation direction & Identical \\
Look/squint angle difference & $\leq 2$ \textdegree \\
\hline
\end{tabular}
\end{table}

\begin{table}[H]
\centering
\caption{Temporal Constraints}
\begin{tabular}{ll}
\hline
\textbf{Interval Type} & \textbf{Time Range} \\
\hline
Short-term & 14-105 days \\
Medium-term & $\sim$1 year $\pm$ 45 days \\
Long-term & $\sim$2 years $\pm$ 45 days \\
\hline
\end{tabular}
\end{table}

\begin{table}[H]
\centering
\caption{Optical Image Requirements}
\begin{tabular}{ll}
\hline
\textbf{Parameter} & \textbf{Requirement} \\
\hline
Temporal proximity to SAR & Within 5 days \\
Cloud coverage & $\geq$ 90\% cloud-free \\
Off-nadir angle & $< 30$ \textdegree \\
Intersection over Union (IoU) with SAR footprint & $\geq 60\%$ \\
\hline
\end{tabular}
\end{table}

\begin{table}[H]
\centering
\caption{SkyCap Dataset Composition}
\begin{tabular}{ll}
\hline
\textbf{Metric} & \textbf{Value} \\
\hline
Minimum overlap area & $\geq$ 15 km² \\
Number of quartets (after deduplication) & 19 \\
Geographic coverage & Eastern Europe, the Middle East, parts of Asia \\
Biomes covered & Tropical, desert, temperate \\
Co-registration accuracy (MAE) & 5 pixels (2.5m) \\
Keypoints used for registration & 30-50 \\
Tile size & 512 $\times$ 512 pixels \\
\textbf{Total image-pairs (pretraining)} & \textbf{68,000} \\
\textbf{Annotated image-pair samples with changes (change detection)} & \textbf{3,484} \\
\hline
\end{tabular}
\end{table}

\newpage

\section{Training details}
\label{app:training}

All models are fully end-to-end fine-tuned on images of size 512x512 pixels for 50 epochs with AdamW optimizer. The SAR models are trained with a learning rate of $1e-5$, weight decay of 0.05, batch size of 64, a cosine annealing scheduler with 20\% linear warmup, and combined class-weighted Dice + weighted cross-entropy loss. The random initialized U-Net decoder receives an LR multiplier of 10. The optical models follow the same protocol, except for a batch size of 16, learning rate of $3e-5$ and weight decay of 0.1. We follow a standard 70\%/10\%/20\% train/val/test split with random patch-level splits within each scene quartet, due to the small number and diversity of SkyCap quartets making a representative scene-level split infeasible. All models were trained on a single NVIDIA L40S with 48GB of VRAM, and training takes between 4.5 and 6.5 hours, depending on the backbone.

\end{document}